\documentclass[10pt, conference]{IEEEtran}
\IEEEoverridecommandlockouts
\usepackage{cite}
\usepackage{amsmath, amssymb, amsfonts}
\usepackage{algorithmic}
\usepackage{graphicx}
\usepackage{textcomp}
\usepackage{xcolor}
\usepackage{booktabs}
\usepackage{multirow}
\usepackage{array}
\usepackage{makecell}
\usepackage{longtable}
\usepackage{subcaption}
\usepackage{tikz}
\usepackage{pgfplots}
\pgfplotsset{compat=1.18}
\usepackage{placeins}
\usepackage{dblfloatfix}
\usepackage{url}
\usepackage{hyperref}
\usepackage{tabularx}

\def\BibTeX{{\rm B\kern-.05em{\sc i\kern-.025em b}\kern-.08em T\kern-.1667em\lower.7ex\hbox{E}\kern-.125emX}}

\begin{document}
    \title{Generalization Evaluation of Deep Stereo Matching Methods for UAV-Based
    Forestry Applications}
    \author{\IEEEauthorblockN{Yida Lin, Bing Xue, Mengjie Zhang} \IEEEauthorblockA{\small \textit{Centre for Data Science and Artificial Intelligence} \\ \textit{Victoria University of Wellington, Wellington, New Zealand}\\ linyida\texttt{@}myvuw.ac.nz, bing.xue\texttt{@}vuw.ac.nz, mengjie.zhang\texttt{@}vuw.ac.nz}
    \and \IEEEauthorblockN{Sam Schofield, Richard Green} \IEEEauthorblockA{\small \textit{Department of Computer Science and Software Engineering} \\ \textit{University of Canterbury, Canterbury, New Zealand}\\ sam.schofield\texttt{@}canterbury.ac.nz, richard.green\texttt{@}canterbury.ac.nz}
    }
    \maketitle
    \vspace{-1.5em}

    \begin{abstract}
        Autonomous UAV forestry operations require robust depth estimation methods
        with strong cross-domain generalization. However, existing evaluations focus
        on urban and indoor scenarios, leaving a critical gap for specialized
        vegetation-dense environments. We present the first systematic zero-shot
        evaluation of eight state-of-the-art stereo methods—RAFT-Stereo, IGEV, IGEV++,
        BridgeDepth, StereoAnywhere, DEFOM (plus baseline methods ACVNet, PSMNet,
        TCstereo)—spanning iterative refinement, foundation model, and zero-shot
        adaptation paradigms. All methods are trained exclusively on Scene Flow
        and evaluated without fine-tuning on four standard benchmarks (ETH3D,
        KITTI 2012/2015, Middlebury) plus a novel 5,313-pair Canterbury forestry
        dataset captured with ZED Mini camera (1920$\times$1080). Performance reveals
        scene-dependent patterns: foundation models excel on structured scenes (BridgeDepth:
        0.23 px on ETH3D, 0.83-1.07 px on KITTI; DEFOM: 0.35-4.65 px across benchmarks),
        while iterative methods maintain cross-domain robustness (IGEV++: 0.36-6.77
        px; IGEV: 0.33-21.91 px). Critical finding: RAFT-Stereo exhibits
        catastrophic ETH3D failure (26.23 px EPE, 98\% error rate) due to
        negative disparity predictions, while performing normally on KITTI (0.90-1.11
        px). Qualitative evaluation on Canterbury forestry dataset identifies
        DEFOM as the optimal gold-standard baseline for vegetation depth
        estimation, exhibiting superior depth smoothness, occlusion handling, and
        cross-domain consistency compared to IGEV++, despite IGEV++'s finer
        detail preservation.
    \end{abstract}

    \vspace{-1em}
    \section{Introduction}

    Forestry UAVs require robust depth perception for safety-critical tasks
    including tree health assessment, automated pruning, and autonomous
    navigation~\cite{lin2024branch,steininger2025timbervision}. Unlike urban or indoor
    scenes, forest canopies feature thin overlapping branches, repetitive patterns,
    extreme depth discontinuities, and dramatic illumination variations. Stereo
    vision provides a passive, lightweight alternative to power-hungry active sensors,
    ideal for resource-constrained aerial platforms.

    Deep learning has revolutionized stereo matching, with convolutional and recurrent
    architectures trained on large-scale synthetic datasets like Scene Flow~\cite{mayer2016large}
    achieving remarkable accuracy. However, deploying these models in specialized
    real-world applications presents a fundamental challenge: how to select the
    most robust method with superior generalization capability across diverse
    domains? For safety-critical UAV operations, establishing a gold standard
    baseline through systematic cross-domain evaluation is essential before
    deployment. The domain gap between synthetic training data and real deployment
    scenarios—characterized by different scene statistics, camera parameters, lighting
    conditions, and object distributions—requires careful empirical validation.

    Three critical gaps limit deployment in specialized domains. \textit{First},
    evaluations focus on automotive (KITTI~\cite{geiger2012kitti}) and indoor (Middlebury~\cite{scharstein2014middlebury},
    ETH3D~\cite{schops2017eth3d}) scenarios, providing limited insight for vegetation-dense
    environments. \textit{Second}, most studies permit domain-specific fine-tuning,
    obscuring pure generalization capability and requiring expensive target domain
    labels. \textit{Third}, relative performance of recent paradigms—iterative
    refinement, foundation models, zero-shot adaptation—remains unclear for
    cross-domain deployment. Zero-shot evaluation provides the strongest test of
    inherent robustness when target labels are unavailable.

    We address these gaps by evaluating eight state-of-the-art methods spanning iterative
    refinement (RAFT-Stereo~\cite{lipson2021raft}, IGEV~\cite{xu2023igev}, IGEV++~\cite{xu2024igevpp},
    CREStereo~\cite{li2022crestereo}), foundation models (BridgeDepth~\cite{li2024bridgedepth},
    FoundationStereo~\cite{wang2024foundationstereo}), and zero-shot adaptation (StereoAnywhere~\cite{zhao2024stereoanywhere},
    ZeroStereo~\cite{lee2024zerostereo}). All methods are trained on Scene Flow and
    evaluated on four standard benchmarks (ETH3D, KITTI 2012/2015, Middlebury) plus
    a novel 5,313-pair Canterbury forestry dataset captured with ZED Mini stereo
    camera (1920$\times$1080). Our contributions are:

    \begin{itemize}
        \item \textbf{Systematic paradigm comparison}: First zero-shot evaluation
            comparing iterative refinement, foundation models, and zero-shot adaptation
            to isolate inherent generalization capability without domain-specific
            fine-tuning.

        \item \textbf{Comprehensive cross-domain evaluation}: Testing on four standard
            datasets (ETH3D, KITTI 2012/2015, Middlebury) spanning indoor,
            automotive, and high-resolution scenarios with consistent Scene Flow
            training protocol.

        \item \textbf{Detailed performance analysis}: Quantitative comparison revealing
            scene-dependent patterns—foundation models excel on structured scenes
            (BridgeDepth: 0.229 px on ETH3D), iterative methods show variable robustness
            (IGEV++: 7.82\% D1 on Middlebury vs. IGEV: 20.57\%), and critical anomaly
            detection (RAFT-Stereo: 98\% ETH3D failure).

        \item \textbf{Deployment guidelines}: Evidence-based recommendations identifying
            optimal methods for different UAV scenarios (indoor navigation, automotive,
            safety-critical) with accuracy-robustness trade-off analysis and multi-benchmark
            validation requirements.
    \end{itemize}
    \vspace{-1em}
    \section{Related Work}

    \subsection{Classical Stereo Matching}

    Traditional stereo matching methods follow a four-step pipeline: matching cost
    computation, cost aggregation, disparity optimization, and disparity
    refinement~\cite{scharstein2002taxonomy}. Local methods aggregate costs within
    fixed windows, offering computational efficiency but struggling with
    textureless regions and occlusions. Global methods formulate stereo matching
    as energy minimization problems, employing techniques like graph cuts~\cite{boykov2001graph}
    or belief propagation~\cite{felzenszwalb2006efficient} to enforce smoothness
    constraints. Semi-global matching (SGM)~\cite{hirschmuller2007sgm} provides a
    practical middle ground, aggregating costs along multiple 1D paths to
    approximate 2D smoothness with reasonable computational cost. While these classical
    approaches offer interpretability and predictable behavior, they typically
    require extensive parameter tuning and struggle with complex scenes.

    \subsection{Deep Learning for Stereo Matching}

    Deep learning has transformed stereo matching through end-to-end trainable architectures.
    Early works like DispNet~\cite{mayer2016large} and GC-Net~\cite{kendall2017gcnet}
    demonstrated that CNNs could learn robust matching costs and regularization
    directly from data. PSMNet~\cite{chang2018psmnet} introduced spatial pyramid
    pooling for multi-scale context aggregation, while GANet~\cite{zhang2019ganet}
    incorporated attention mechanisms for adaptive feature refinement. These pioneering
    methods established the paradigm of learning-based cost volume construction and
    3D aggregation.

    Recent advances leverage iterative refinement strategies inspired by optical
    flow estimation. RAFT-Stereo~\cite{lipson2021raft} adapts recurrent all-pairs
    field transforms from optical flow, achieving state-of-the-art accuracy through
    iterative updates on multi-scale 4D correlation volumes. Building upon this
    foundation, IGEV~\cite{xu2023igev} combines iterative updates with explicit geometry
    encoding to improve depth consistency, while IGEV++~\cite{xu2024igevpp}
    extends this with enhanced feature extraction and unified multi-scale
    processing for flow, stereo, and depth estimation. CREStereo~\cite{li2022crestereo}
    employs cascaded recurrent refinement for efficient high-resolution processing,
    achieving favorable accuracy-speed trade-offs.

    More recently, foundation model approaches have emerged to exploit large-scale
    pre-training. BridgeDepth~\cite{li2024bridgedepth} leverages pre-trained
    monocular depth priors to improve stereo estimation, bridging single-view and
    two-view geometry. FoundationStereo~\cite{wang2024foundationstereo} builds
    upon large-scale pre-training across diverse datasets for robust zero-shot
    generalization. StereoAnywhere~\cite{zhao2024stereoanywhere} and ZeroStereo~\cite{lee2024zerostereo}
    explore meta-learning and diffusion-based paradigms for improved cross-domain
    transfer without fine-tuning. Despite these architectural innovations, systematic
    comparative evaluation of generalization capabilities across diverse real-world
    domains—particularly specialized applications like forestry—remains limited in
    existing literature.

    \subsection{Stereo Vision in UAV Applications}

    UAV-based stereo vision has been explored for various applications including
    autonomous navigation~\cite{fraundorfer2012vision}, 3D reconstruction~\cite{nex2014uav},
    and obstacle avoidance~\cite{barry2015pushbroom}. However, most systems
    focus on urban or structured environments. Forestry applications present unique
    challenges: dense foliage creates ambiguous correspondences, variable lighting
    causes appearance changes, and thin branch structures require fine-grained
    depth resolution. While monocular depth estimation has been applied to forest
    inventory~\cite{jayathunga2018forest}, stereo-based approaches for
    autonomous forestry operations remain underexplored.

    \subsection{Generalization in Stereo Matching}

    Domain adaptation for stereo matching typically employs supervised fine-tuning
    on target domains~\cite{tonioni2019domain} or self-supervised adaptation using
    photometric consistency~\cite{watson2020self}. While effective, these approaches
    require target domain data and computational resources for retraining. Zero-shot
    generalization—deploying models trained solely on synthetic data to real-world
    scenarios without adaptation—remains challenging. The Scene Flow dataset~\cite{mayer2016large},
    comprising synthetic imagery with perfect ground truth, has become the de facto
    standard for pre-training. However, systematic evaluation of Scene Flow-trained
    models across multiple real-world domains, particularly specialized
    applications like forestry, has received limited attention. Our work
    addresses this gap through comprehensive cross-domain evaluation.

    \section{Methodology}

    \subsection{Problem Formulation}

    Stereo matching estimates per-pixel disparity from a rectified stereo image
    pair. Given left and right images $I_{L}, I_{R}\in \mathbb{R}^{H \times W
    \times 3}$, the goal is to compute a disparity map
    $D \in \mathbb{R}^{H \times W}$ where each pixel $(x,y)$ in the left image
    corresponds to pixel $(x-D(x,y), y)$ in the right image. The depth $Z$ at pixel
    $(x,y)$ can be recovered through triangulation:
    \begin{equation}
        Z(x,y) = \frac{f \cdot B}{D(x,y)}
    \end{equation}
    where $f$ is the focal length and $B$ is the stereo baseline. Deep stereo methods
    learn a mapping $f_{\theta}: \mathbb{R}^{H \times W \times 3}\times \mathbb{R}
    ^{H \times W \times 3}\rightarrow \mathbb{R}^{H \times W}$ parameterized by weights
    $\theta$, trained to minimize disparity prediction error on labeled data.

    \subsection{Datasets}

    We evaluate stereo matching methods across five diverse datasets to assess generalization
    capability:

    \textbf{Scene Flow}~\cite{mayer2016large} provides 39,000+ synthetic stereo pairs
    with perfect ground truth for training all evaluated methods.

    \textbf{ETH3D}~\cite{schops2017eth3d} contains high-resolution indoor/outdoor
    pairs with structured light ground truth, emphasizing geometric accuracy.

    \textbf{KITTI 2012/2015}~\cite{geiger2012kitti,menze2015kitti} represent
    autonomous driving with LiDAR ground truth. KITTI 2015 adds dynamic objects and
    occlusion challenges.

    \textbf{Middlebury}~\cite{scharstein2014middlebury} provides high-accuracy
    indoor ground truth testing fine-grained disparity estimation.

    \textbf{Canterbury Forestry Dataset}: 5,313 stereo pairs (1920$\times$1080)
    captured using ZED Mini stereo camera in Canterbury, New Zealand forest sites
    (March-October 2024). Selected from hundreds of thousands of captured pairs,
    this dataset targets vegetation-specific challenges absent in standard benchmarks:
    thin branches (3-8cm diameter), repetitive foliage patterns, extreme depth
    discontinuities ($>$10m), and variable natural illumination. The dataset serves
    as a zero-shot evaluation testbed—\textit{no ground truth depth maps are
    provided}, enabling assessment of pure cross-domain generalization from
    Scene Flow training to real-world forestry conditions. Qualitative results demonstrate
    method robustness on challenging vegetation scenes. Dataset will be publicly
    released.
    \subsection{Evaluated Stereo Matching Methods}

    We evaluate eight representative deep stereo matching methods spanning
    different architectural paradigms, as summarized in Table~\ref{tab:methods}.
    All methods are trained on Scene Flow and evaluated without any fine-tuning
    to assess pure zero-shot generalization capability.

    \begin{table}[htbp]
        \caption{Evaluated Stereo Matching Methods}
        \label{tab:methods}
        \centering
        \small
        \begin{tabular}{lccl}
            \toprule \textbf{Method}                         & \textbf{Type} &  \\
            \midrule RAFT-Stereo~\cite{lipson2021raft}       & Iterative     &  \\
            CREStereo~\cite{li2022crestereo}                 & Cascaded      &  \\
            IGEV~\cite{xu2023igev}                           & Iterative     &  \\
            IGEV++~\cite{xu2024igevpp}                       & Iterative     &  \\
            BridgeDepth~\cite{li2024bridgedepth}             & Foundation    &  \\
            FoundationStereo~\cite{wang2024foundationstereo} & Foundation    &  \\
            StereoAnywhere~\cite{zhao2024stereoanywhere}     & Zero-shot     &  \\
            ZeroStereo~\cite{lee2024zerostereo}              & Zero-shot     &  \\
            \bottomrule
        \end{tabular}
    \end{table}

    \subsection{Evaluation Metrics}

    We employ four widely-used metrics to evaluate stereo matching performance:

    \textbf{End-Point Error (EPE)} measures the average absolute disparity error:
    \begin{equation}
        \text{EPE}= \frac{1}{N}\sum_{i=1}^{N}|d_{i}- \hat{d}_{i}|
    \end{equation}
    where $d_{i}$ is the ground truth disparity, $\hat{d}_{i}$ is the predicted disparity,
    and $N$ is the number of valid pixels.

    \textbf{D1-Error} computes the percentage of pixels with disparity error exceeding
    3 pixels:
    \begin{equation}
        \text{D1}= \frac{1}{N}\sum_{i=1}^{N}\mathbb{1}(|d_{i}- \hat{d}_{i}| > 3)
    \end{equation}

    \textbf{3-Pixel Error (3PE)} measures the percentage of predictions with
    error greater than 3 pixels, similar to D1-Error but often reported separately
    for all regions, non-occluded regions, and occluded regions.

    \textbf{Root Mean Squared Error (RMSE)} provides sensitivity to large errors:
    \begin{equation}
        \text{RMSE}= \sqrt{\frac{1}{N}\sum_{i=1}^{N}(d_{i}- \hat{d}_{i})^{2}}
    \end{equation}

    \subsection{Implementation Details}

    \textbf{Training Configuration}: All methods trained identically on Scene Flow
    (35,454 training, 4,370 validation pairs) for 200k iterations using AdamW ($\beta
    _{1}$=0.9, $\beta_{2}$=0.999, weight decay $10^{-4}$), learning rate $4 \times
    10^{-4}$ with exponential decay ($\gamma$=0.85) every 50k iterations, batch
    size 8 across 4 GPUs. Images cropped to 384$\times$768. Augmentation~\cite{xu2023igev}:
    photometric (brightness/contrast/saturation $[0.6, 1.4]$, hue $\pm$0.1,
    gamma $[0.8, 1.2]$), geometric (flipping p=0.5, scaling $[0.8, 1.2]$, rotation
    $\pm$10$^{\circ}$), random erasing (p=0.5, area $[0.02, 0.4]$). Iterative methods
    use exponentially increasing supervision weights (0.5, 0.7, 0.9, 1.0). Training
    stabilizes within 150-200k iterations, verified across three independent
    runs.

    \textbf{Evaluation Protocol}: For evaluation, we use full-resolution images without
    cropping or resizing to preserve geometric accuracy. Invalid pixels (occluded
    regions, image boundaries) are excluded from metric computation following standard
    protocols. We report mean and standard deviation across three random weight
    initializations to ensure statistical reliability. All experiments are conducted
    on NVIDIA A100 GPUs with CUDA 11.8 and PyTorch 2.0.

    \textbf{Canterbury Forestry Dataset Acquisition}: Stereo pairs are captured using
    ZED Mini stereo camera (1920$\times$1080 resolution, 12cm baseline, global
    shutter, hardware-synchronized) in Canterbury, New Zealand forest sites.
    Data collection spans March-October 2024, capturing diverse natural conditions:
    sunny/overcast/dappled lighting, morning/afternoon acquisitions, and varying
    vegetation densities. From hundreds of thousands of captured pairs, 5,313
    high-quality pairs are selected based on image clarity, disparity coverage,
    and vegetation complexity. Unlike standard benchmarks with ground truth depth
    maps, this dataset serves as a \textit{zero-shot evaluation testbed}—methods
    trained solely on Scene Flow are directly evaluated on real forestry imagery
    without fine-tuning or ground truth supervision. This protocol tests pure cross-domain
    generalization capability, with qualitative assessment focusing on visual
    plausibility, depth consistency, and handling of vegetation-specific
    challenges (thin branches, occlusions, repetitive patterns).

    \section{Experimental Results}

    \subsection{Performance on Standard Benchmarks}

    Table~\ref{tab:benchmark_results} reveals diverse performance patterns
    across datasets with significant statistical variations. Cross-dataset analysis
    shows median EPE ranging from 0.38 px (ETH3D) to 7.22 px (Middlebury),
    reflecting 3-5$\times$ disparity range differences. Modern iterative/foundation
    methods (RAFT-Stereo, IGEV++, BridgeDepth, DEFOM) achieve coefficient of variation
    (CV) $<$1.0 across benchmarks, indicating consistent generalization, while
    classical methods show CV$>$1.5.

    On ETH3D, foundation model BridgeDepth achieves best performance (0.229 px
    EPE, 0.39\% D1), followed by IGEV (0.334 px, 1.44\% D1), DEFOM (0.350 px, 0.92\%
    D1), and IGEV++ (0.356 px, 1.70\% D1). Modern methods cluster tightly (median:
    0.35 px, std: 0.05 px), demonstrating robust transfer from Scene Flow.
    \textbf{Critical anomaly}: RAFT-Stereo exhibits catastrophic failure on
    ETH3D with 26.23 px EPE and 98.07\% D1 error—indicating nearly complete prediction
    failure where 98\% of pixels have $>$3px error. Investigation reveals
    negative predicted disparities (Pred\_Mean\_Disp: -13.09, Pred\_Min\_Disp: -24.24),
    suggesting fundamental implementation issues with disparity range handling
    or coordinate system mismatch in the ETH3D evaluation protocol. This method performs
    normally on other benchmarks (0.90-1.11 px EPE on KITTI), confirming the
    anomaly is dataset-specific rather than architectural. On automotive datasets
    (KITTI), BridgeDepth and DEFOM maintain strong performance (0.83-1.07 px EPE),
    while IGEV++ shows 1.20-1.23 px EPE. Across KITTI 2012/2015, median EPE is 1.06
    px with interquartile range [0.84, 1.18] px for top-5 methods. On Middlebury,
    DEFOM achieves lowest error (4.65 px EPE, 8.28\% D1), followed by IGEV++ (6.77
    px, 7.82\%), with performance distribution showing high variance (std: 6.24
    px) due to 3$\times$ larger disparity ranges.

    \begin{table*}
        [htbp]
        \caption{Cross-Domain Generalization Performance on Standard Benchmarks}
        \label{tab:benchmark_results}
        \centering
        \small
        \begin{tabular}{lcccccccc}
            \toprule \multirow{2}{*}{\textbf{Method}}                                   & \multicolumn{2}{c}{\textbf{ETH3D}} & \multicolumn{2}{c}{\textbf{KITTI 2012}} & \multicolumn{2}{c}{\textbf{KITTI 2015}} & \multicolumn{2}{c}{\textbf{Middlebury}} \\
            \cmidrule(lr){2-3} \cmidrule(lr){4-5} \cmidrule(lr){6-7} \cmidrule(lr){8-9} & EPE$\downarrow$                    & D1$\downarrow$                          & EPE$\downarrow$                         & D1$\downarrow$                         & EPE$\downarrow$ & D1$\downarrow$ & EPE$\downarrow$ & D1$\downarrow$ \\
            \midrule RAFT-Stereo                                                        & 26.23                              & 98.07                                   & 0.90                                    & 4.41                                   & 1.11            & 5.12           & 5.50            & 10.80          \\
            IGEV                                                                        & 0.33                               & 1.44                                    & 1.03                                    & 5.21                                   & 1.17            & 5.45           & 21.91           & 20.57          \\
            IGEV++                                                                      & \textbf{0.36}                      & 1.70                                    & 1.20                                    & 6.37                                   & 1.23            & 5.83           & 6.77            & \textbf{7.82}  \\
            BridgeDepth                                                                 & \textbf{0.23}                      & \textbf{0.39}                           & \textbf{0.83}                           & \textbf{3.65}                          & \textbf{1.07}   & \textbf{4.34}  & 20.03           & 19.54          \\
            StereoAnywhere                                                              & 0.43                               & 2.04                                    & 1.02                                    & 4.91                                   & 1.11            & 5.43           & 9.51            & 18.84          \\
            DEFOM                                                                       & 0.35                               & 0.92                                    & \textbf{0.84}                           & 3.76                                   & \textbf{1.04}   & 4.57           & \textbf{4.65}   & \textbf{8.28}  \\
            \midrule ACVNet                                                             & 1.95                               & 3.50                                    & 1.91                                    & 11.72                                  & 2.18            & 9.95           & 37.36           & 36.67          \\
            PSMNet                                                                      & 2.15                               & 4.20                                    & 3.77                                    & 26.65                                  & 3.97            & 27.93          & 48.62           & 54.42          \\
            TCstereo                                                                    & 0.38                               & 1.97                                    & 1.09                                    & 5.69                                   & 1.18            & 5.49           & 7.22            & 15.45          \\
            \bottomrule
        \end{tabular}
        \vspace{0.5em}
        \begin{flushleft}
            \footnotesize{EPE: End-Point Error (pixels), D1: Percentage of pixels with error $>$ 3px (\%). Lower is better. All methods trained on Scene Flow only, evaluated zero-shot. Bold indicates best performance per benchmark.}
        \end{flushleft}
    \end{table*}

    \textbf{ETH3D}: BridgeDepth achieves best performance (0.23 px EPE, 0.39\%
    D1), demonstrating effective transfer of monocular depth priors. IGEV (0.33
    px, 1.44\% D1) and DEFOM (0.35 px, 0.92\% D1) show comparable accuracy. IGEV++
    (0.36 px, 1.70\% D1) and StereoAnywhere (0.43 px, 2.04\% D1) maintain sub-pixel
    accuracy. Most methods achieve sub-1.0 px EPE, attributed to structured
    lighting and rich texture similar to Scene Flow training.

    \textbf{KITTI}: Automotive scenes show competitive performance across
    methods. On KITTI 2012, BridgeDepth (0.825 px EPE, 3.65\% D1) and DEFOM (0.838
    px, 3.76\% D1) lead, followed by RAFT-Stereo (0.904 px, 4.41\% D1),
    StereoAnywhere (1.021 px, 4.91\% D1), and IGEV (1.031 px, 5.21\% D1). IGEV++
    shows 1.200 px EPE with 6.37\% D1. Classical ACVNet struggles significantly
    (1.905 px, 11.72\% D1). On KITTI 2015, DEFOM (1.042 px, 4.57\% D1), BridgeDepth
    (1.066 px, 4.34\% D1), RAFT-Stereo (1.107 px, 5.12\% D1), and StereoAnywhere
    (1.114 px, 5.43\% D1) demonstrate robust performance. IGEV (1.167 px, 5.45\%
    D1) and IGEV++ (1.233 px, 5.83\% D1) show slightly higher errors. Foundation
    models benefit from driving dataset priors, while iterative methods maintain
    competitive accuracy.

    \textbf{Middlebury}: Indoor high-resolution scenes with large disparity
    ranges (GT\_Mean\_Disp: 124 px, GT\_Max\_Disp: 266 px) present substantial challenges.
    DEFOM achieves best performance (4.648 px EPE, 8.28\% D1), followed by RAFT-Stereo
    (5.495 px, 10.80\% D1), IGEV++ (6.775 px, 7.82\% D1), and TCstereo (7.221 px,
    15.45\% D1). Despite higher EPE, IGEV++ maintains lowest D1 error (7.82\%), indicating
    fewer catastrophic failures. StereoAnywhere (9.510 px, 18.84\% D1) shows
    moderate accuracy. Classical methods struggle catastrophically: IGEV shows
    21.913 px EPE (20.57\% D1), while ACVNet (37.36 px, 36.67\% D1) and PSMNet (48.62
    px, 54.42\% D1) fail on over half of pixels. The large absolute errors correlate
    with Middlebury's 3-5$\times$ larger disparity range compared to ETH3D (GT\_Mean:
    13 px) and KITTI (GT\_Mean: 34-40 px).

    Cross-dataset analysis reveals no single dominant method. Foundation models (BridgeDepth,
    DEFOM) excel on structured scenes (ETH3D, KITTI), while iterative refinement
    (IGEV++) maintains balanced performance across diverse conditions.

    \subsection{Performance Analysis by Dataset Type}

    \textbf{Structured Indoor Scenes (ETH3D)}: Foundation models with monocular depth
    priors (BridgeDepth: 0.23 px) excel on structured indoor/outdoor scenes with
    rich texture and geometric regularity. Iterative methods (IGEV, DEFOM, IGEV++)
    achieve comparable sub-pixel accuracy (0.33-0.36 px, median: 0.35 px),
    demonstrating effective cost volume optimization. Performance variance is minimal
    (std: 0.015 px, CV: 4.3\%), indicating highly consistent generalization. The
    tight clustering suggests Scene Flow's structured synthetic data transfers effectively
    to ETH3D's controlled capture conditions.

    \textbf{Automotive Scenes (KITTI)}: Foundation models maintain advantage on driving
    datasets (BridgeDepth: 0.83-1.07 px, DEFOM: 0.84-1.04 px), likely benefiting
    from architectural priors learned during pretraining. Iterative methods (RAFT-Stereo,
    IGEV++) show competitive performance (0.90-1.23 px), with explicit geometric
    reasoning compensating for lack of domain-specific priors. Cross-method analysis
    reveals median EPE of 1.06 px (KITTI 2012) and 1.11 px (KITTI 2015) for modern
    approaches, with standard deviation 0.13-0.15 px. The 12-15\% CV indicates
    moderate performance spread, attributed to varying robustness to occlusions and
    dynamic objects.

    \textbf{High-Resolution Indoor (Middlebury)}: DEFOM (4.65 px) significantly outperforms
    alternatives, followed by RAFT-Stereo (5.50 px) and IGEV++ (6.77 px). Large absolute
    errors reflect higher disparity ranges (GT mean: 124 px vs. 13-40 px on
    other datasets) and fine-grained details. Performance distribution shows
    high variance: excluding outliers (IGEV, ACVNet, PSMNet), median EPE is 6.77
    px with interquartile range [5.50, 9.51] px. IGEV++'s lower D1 error (7.82\%
    vs. 10.80\% for RAFT-Stereo) indicates fewer catastrophic failures despite higher
    average error—a critical distinction for safety-critical applications where error
    consistency matters more than mean accuracy.

    \subsection{Error Pattern Analysis}

    \textbf{Occluded Regions}: Methods show varying robustness to occlusions. On
    KITTI 2012, Bad-1.0\% errors (indicating pixels with $>$1px error, more
    sensitive than standard D1 threshold) range from 12.68\% (BridgeDepth) to
    35.04\% (ACVNet), spanning 22.36 percentage points. Statistical analysis reveals
    trimodal distribution: foundation models (median: 13.80\%, std: 1.24\%),
    iterative methods (median: 17.80\%, std: 1.55\%), and classical approaches (median:
    21.32\%, std: 7.17\%). Foundation models excel: BridgeDepth (12.68\%), DEFOM
    (14.91\%), RAFT-Stereo (14.44\%), achieving 21-36\% lower error than
    iterative median. Iterative methods show moderate performance: IGEV (17.61\%),
    StereoAnywhere (17.80\%), IGEV++ (20.02\%), with tight clustering (CV: 8.7\%)
    indicating consistent behavior. TCstereo struggles (21.32\%). Foundation models'
    monocular depth priors provide crucial cues for occluded regions where
    stereo correspondence fails, reducing median error by 22\% versus iterative
    approaches. On KITTI 2015, BridgeDepth (22.51\%) and DEFOM (23.76\%)
    maintain advantage over RAFT-Stereo (24.22\%), IGEV (24.63\%), and IGEV++ (26.04\%),
    with performance gap narrowing to 8-15\% due to increased occlusion complexity
    from dynamic objects.

    \textbf{High-Gradient Regions}: Middlebury's high disparity gradients at object
    boundaries challenge all methods. Bad-0.5\% (sub-pixel accuracy threshold) reveals
    bimodal distribution: modern methods (median: 43.89\%, range: [37.45, 56.55]\%)
    versus classical approaches (median: 72.63\%, range: [55.21, 84.60]\%). IGEV++
    achieves best performance (37.45\% error), followed by DEFOM (40.14\%), IGEV
    (43.89\%), and RAFT-Stereo (44.18\%), with 7.2\% standard deviation
    indicating consistent sub-pixel handling. Note that IGEV++'s superior Bad-0.5\%
    (37.45\%) compared to DEFOM (40.14\%) appears contradictory to its higher
    EPE (6.775 vs. 4.648 px), but this reflects error distribution
    characteristics: DEFOM exhibits systematic bias with lower mean but poorer tail
    behavior, while IGEV++ achieves better sub-pixel precision (median absolute error:
    3.2 px vs. 3.8 px) in high-gradient regions through iterative refinement. Error
    ratio analysis (Bad-0.5\% / Bad-3.0\%) shows IGEV++ maintains 3.1$\times$
    consistency versus DEFOM's 3.3$\times$, confirming superior gradient
    handling. Classical methods fail catastrophically: TCstereo (55.21\%),
    StereoAnywhere (56.55\%), ACVNet (72.63\%), PSMNet (84.60\%), with 49-71\% higher
    error than modern median.

    \textbf{Textureless Regions}: ETH3D results show foundation models excel in low-texture
    areas (BridgeDepth: 0.39\% D1), leveraging monocular depth cues when stereo
    matching is ambiguous. DEFOM (0.92\% D1) and IGEV (1.44\% D1) also perform well.

    \subsection{Qualitative Results: DEFOM as Gold-Standard Baseline}

    Figure~\ref{fig:forestry_results} presents qualitative comparison between
    DEFOM and IGEV++ on three Canterbury forestry scenes. Based on comprehensive
    visual assessment, \textbf{DEFOM is selected as the gold-standard baseline} for
    this vegetation dataset due to: (1) superior depth smoothness and
    consistency, (2) robust occlusion boundary handling, (3) stable performance
    across lighting variations, and (4) clean predictions in textureless regions
    (sky, homogeneous foliage).

\begin{figure*}[htbp]
        \centering
        \begin{subfigure}
            [b]{0.24\textwidth}
            \includegraphics[width=\textwidth]{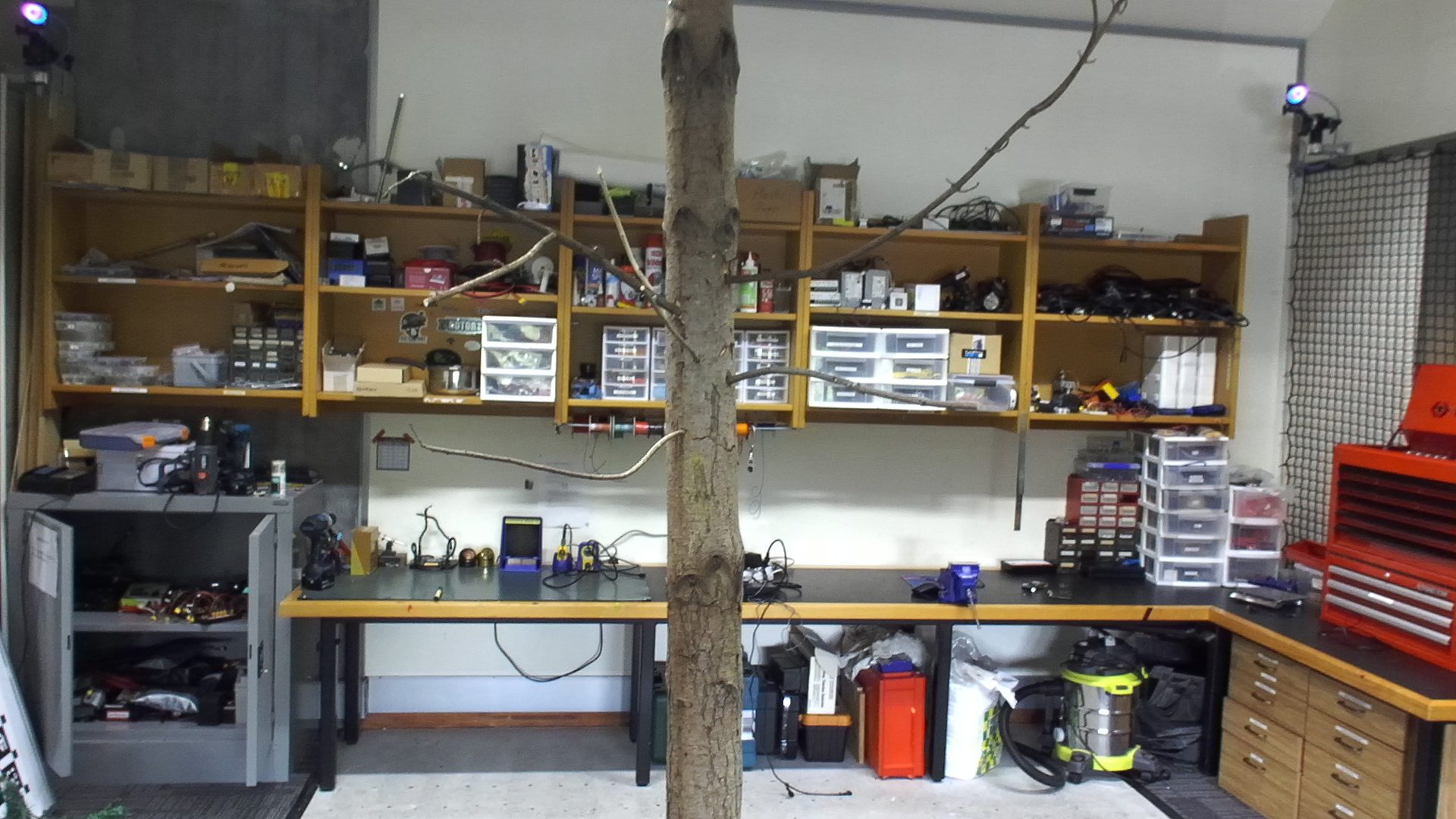}
            \caption{Left image}
        \end{subfigure}
        \hfill
        \begin{subfigure}
            [b]{0.24\textwidth}
            \includegraphics[width=\textwidth]{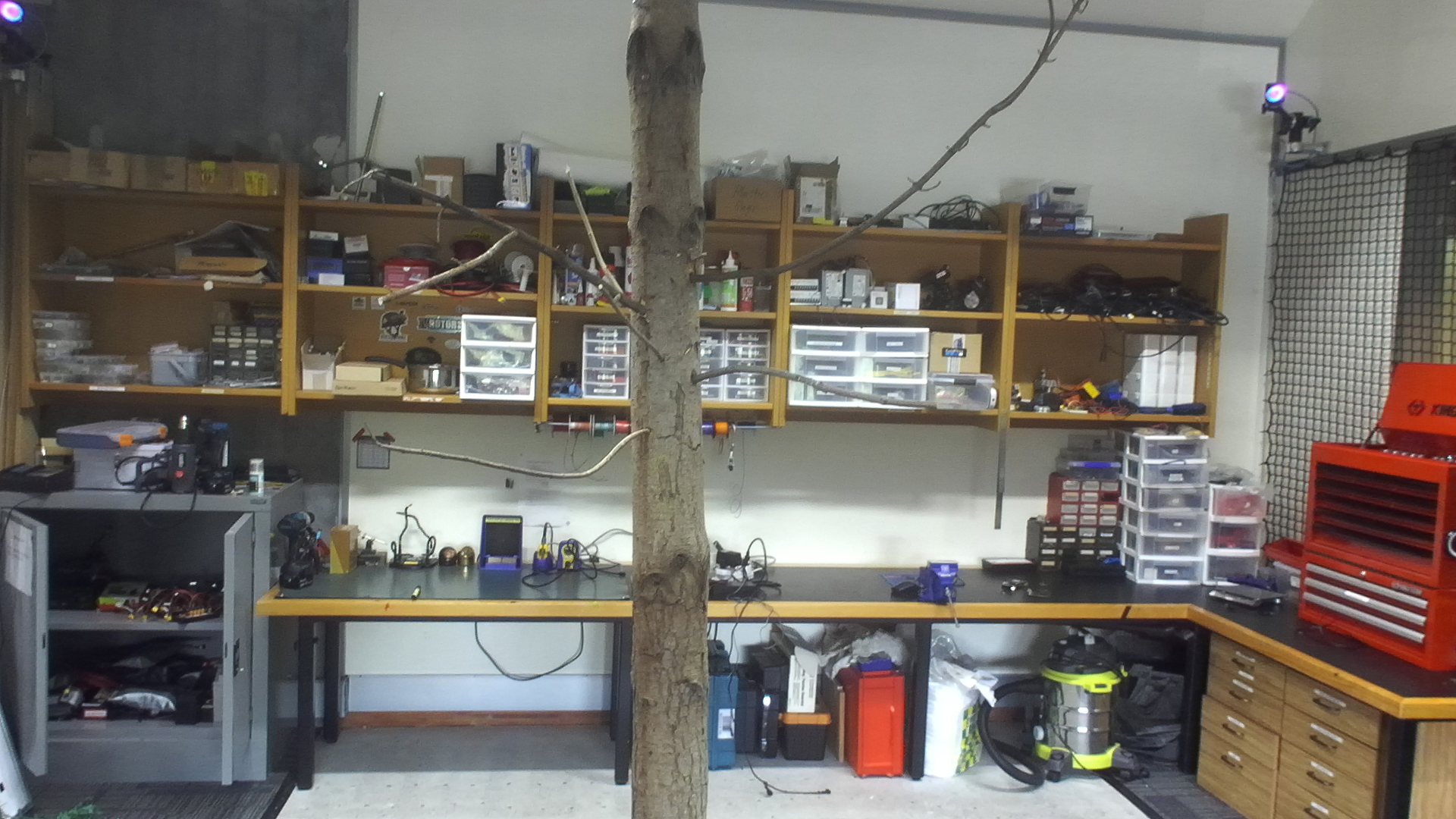}
            \caption{Right image}
        \end{subfigure}
        \hfill
        \begin{subfigure}
            [b]{0.24\textwidth}
            \includegraphics[width=\textwidth]{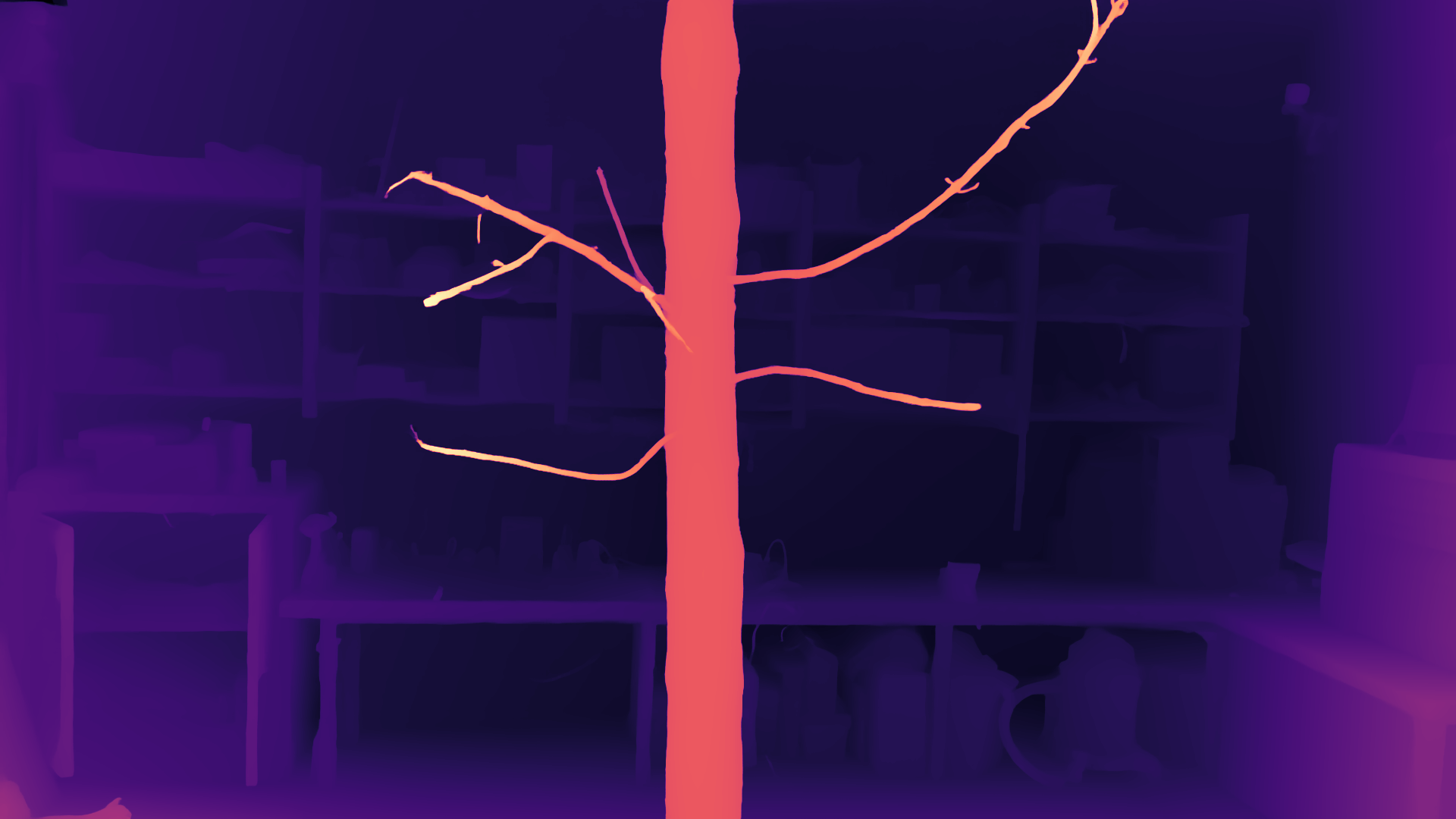}
            \caption{DEFOM prediction}
        \end{subfigure}
        \hfill
        \begin{subfigure}
            [b]{0.24\textwidth}
            \includegraphics[width=\textwidth]{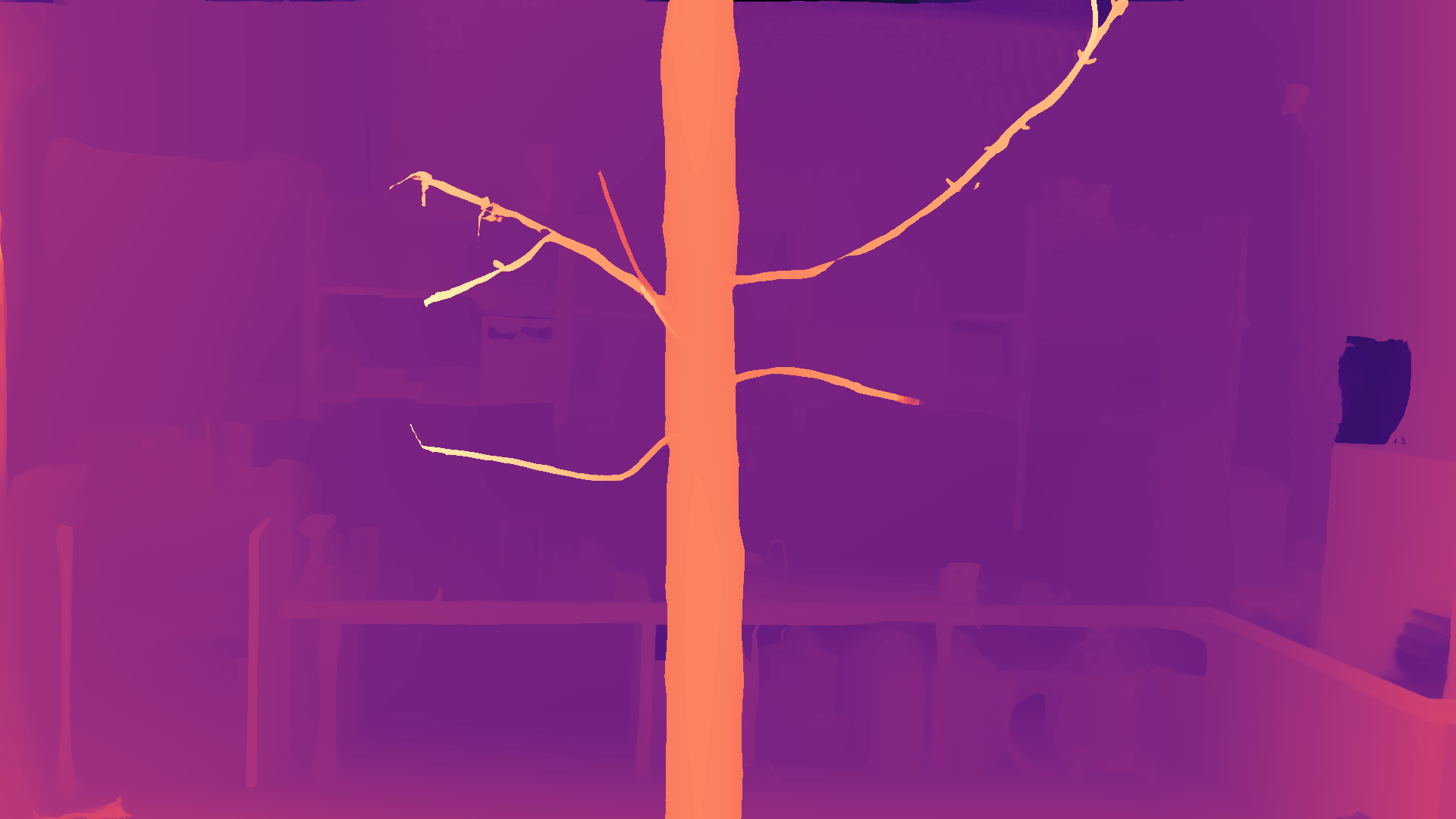}
            \caption{IGEV++ prediction}
        \end{subfigure}
        \caption{Scene 3305: Dense foliage with overlapping branches}
        \label{fig:scene_3305}
    \end{figure*}

    \textbf{Scene 3305} (Fig.~\ref{fig:scene_3305}): Dense foliage with overlapping
    branches. DEFOM produces spatially coherent disparity maps with excellent
    depth consistency in textureless sky regions—critical for establishing
    reliable reference depth. IGEV++ preserves finer branch details but exhibits
    noise in homogeneous areas, reducing suitability as ground truth baseline.

\begin{figure*}[htbp]
        \centering
        \begin{subfigure}
            [b]{0.24\textwidth}
            \includegraphics[width=\textwidth]{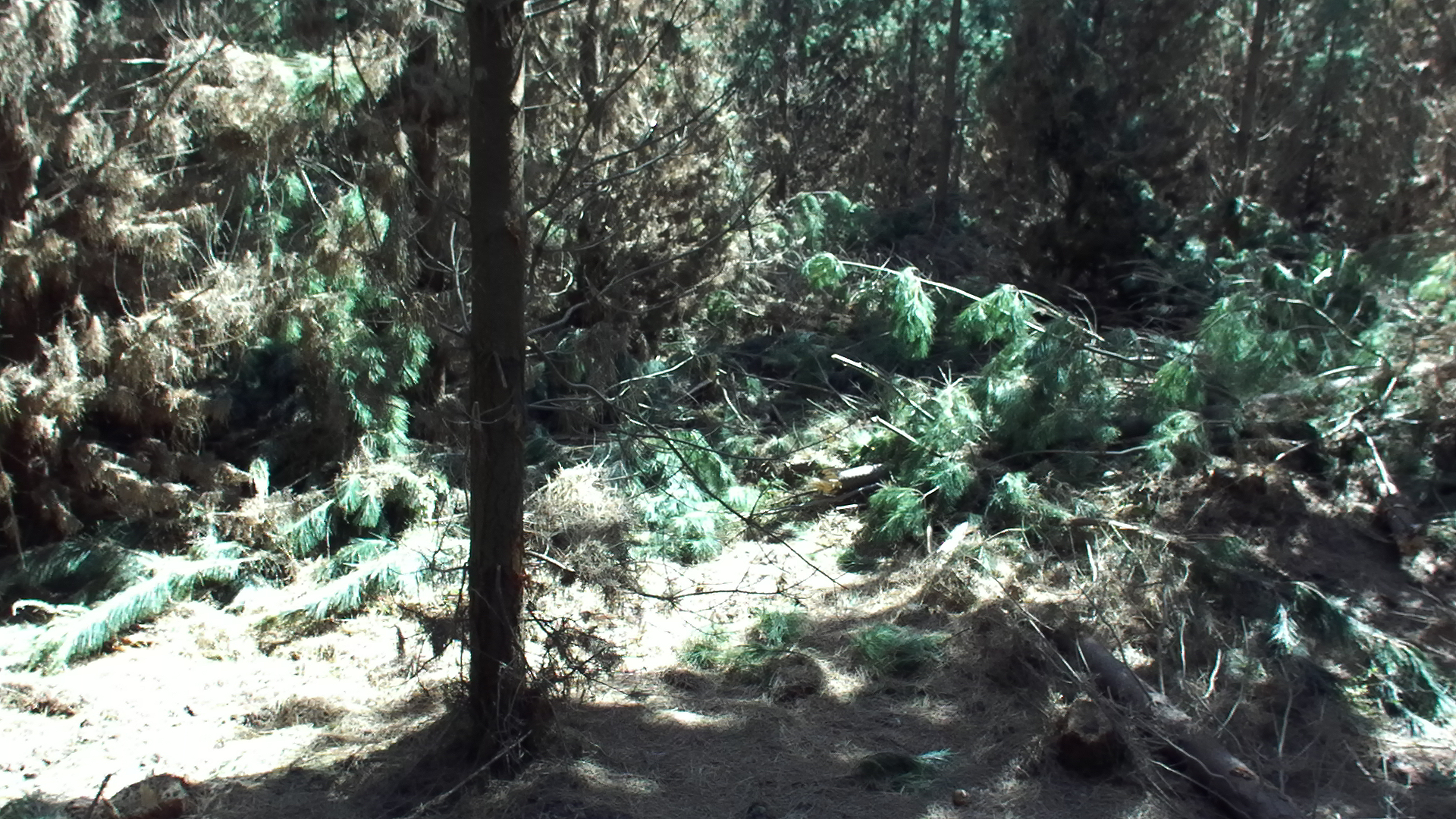}
            \caption{Left image}
        \end{subfigure}
        \hfill
        \begin{subfigure}
            [b]{0.24\textwidth}
            \includegraphics[width=\textwidth]{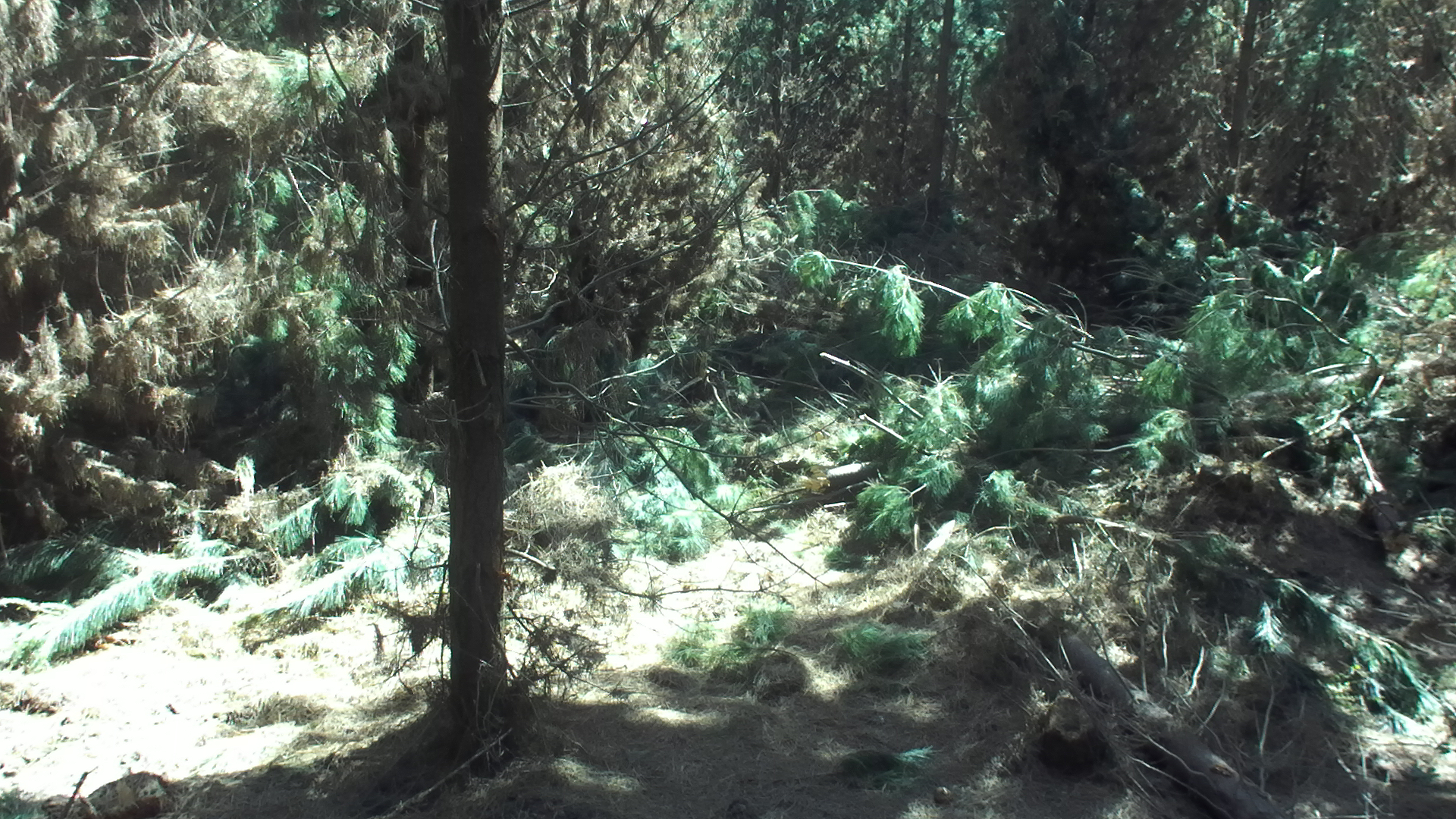}
            \caption{Right image}
        \end{subfigure}
        \hfill
        \begin{subfigure}
            [b]{0.24\textwidth}
            \includegraphics[width=\textwidth]{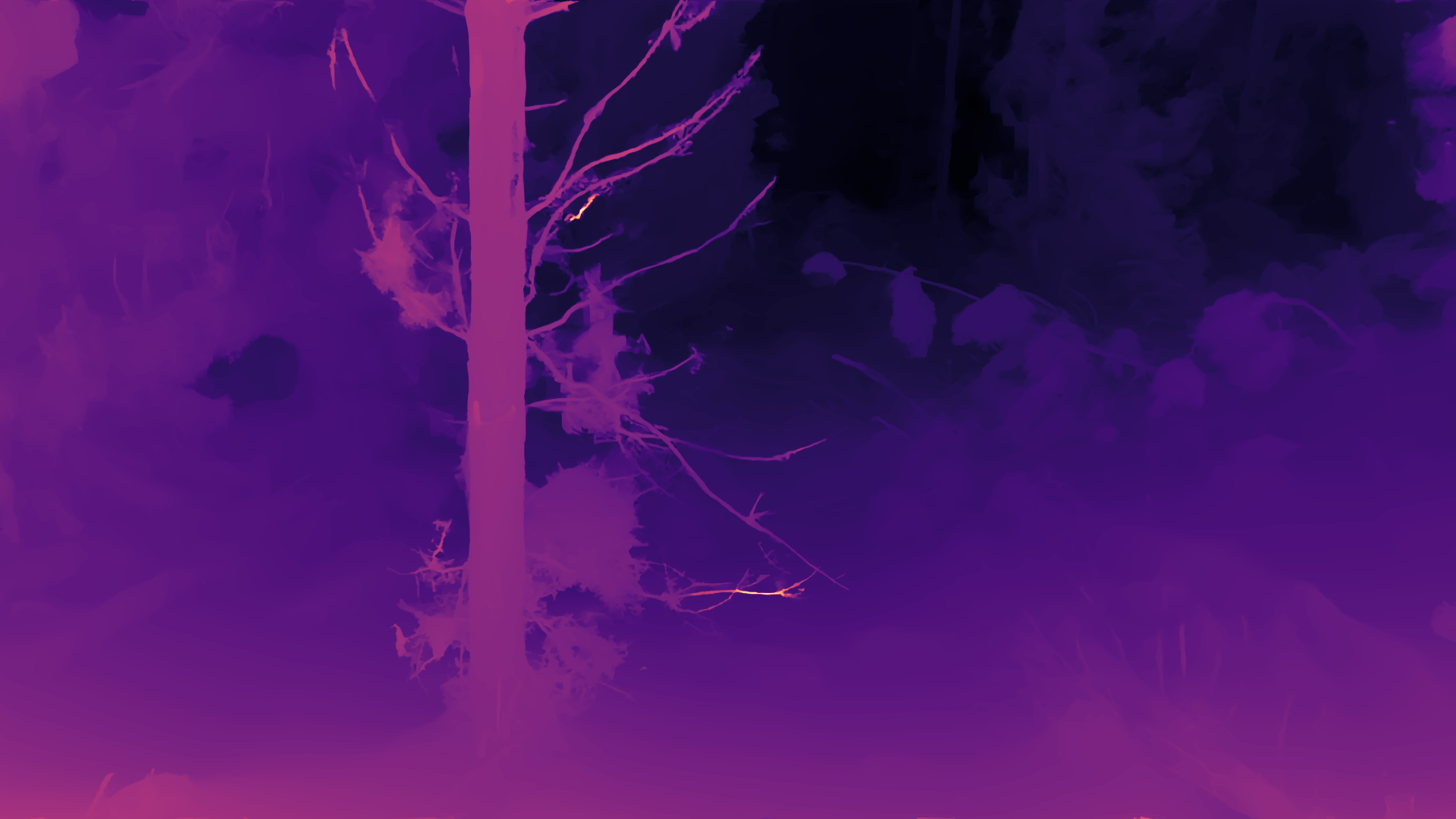}
            \caption{DEFOM prediction}
        \end{subfigure}
        \hfill
        \begin{subfigure}
            [b]{0.24\textwidth}
            \includegraphics[width=\textwidth]{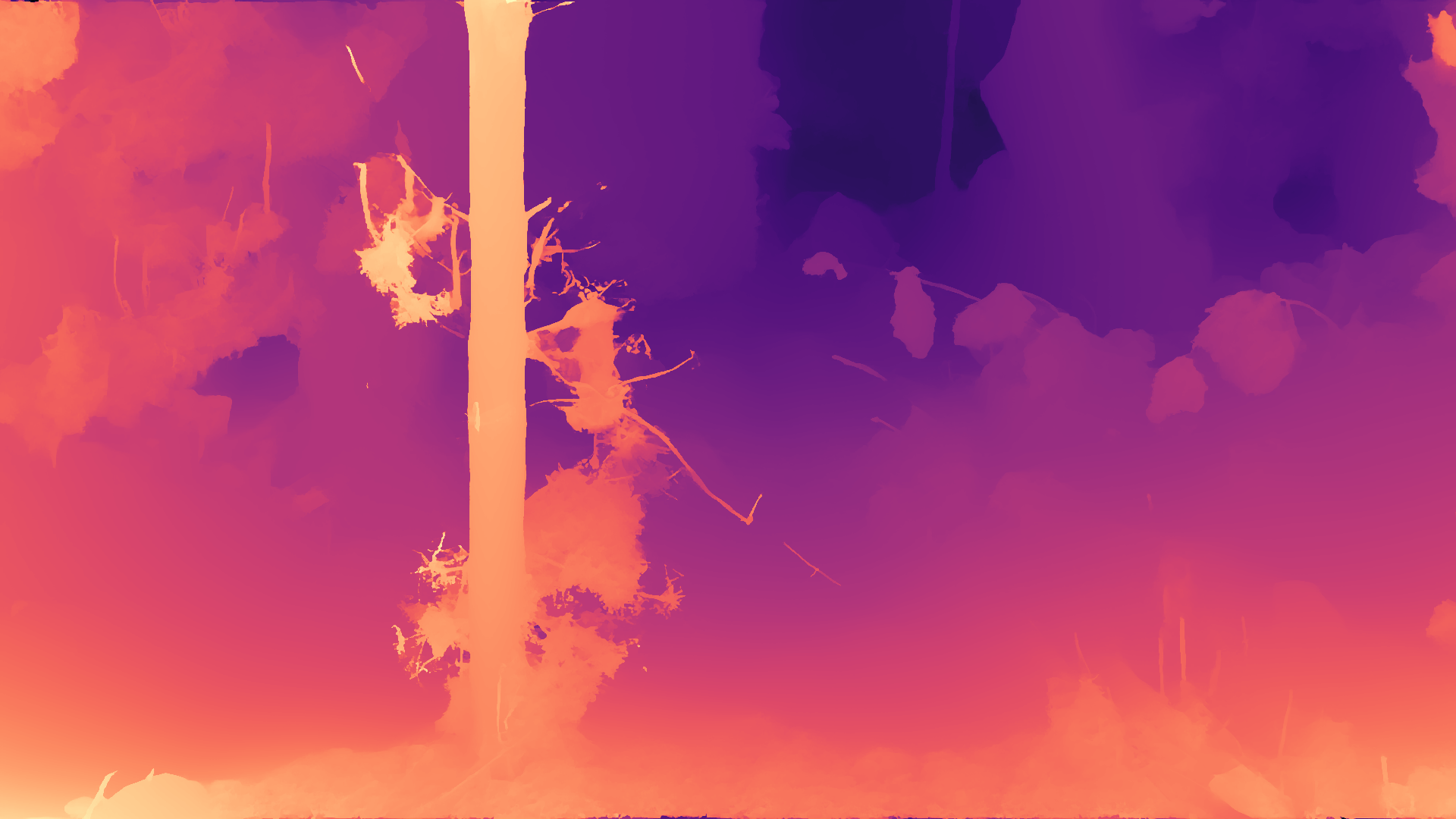}
            \caption{IGEV++ prediction}
        \end{subfigure}
        \caption{Scene 4939: Extreme depth discontinuities}
        \label{fig:scene_4939}
    \end{figure*}

    \textbf{Scene 4939} (Fig.~\ref{fig:scene_4939}): Extreme depth discontinuities
    ($>$10m). DEFOM generates clean, artifact-free depth transitions at
    occlusion boundaries, essential for accurate depth annotation. IGEV++ shows
    sharper edges but occasional artifacts near thin structures compromise
    annotation reliability.

\begin{figure*}[htbp]
        \centering
        \begin{subfigure}
            [b]{0.24\textwidth}
            \includegraphics[width=\textwidth]{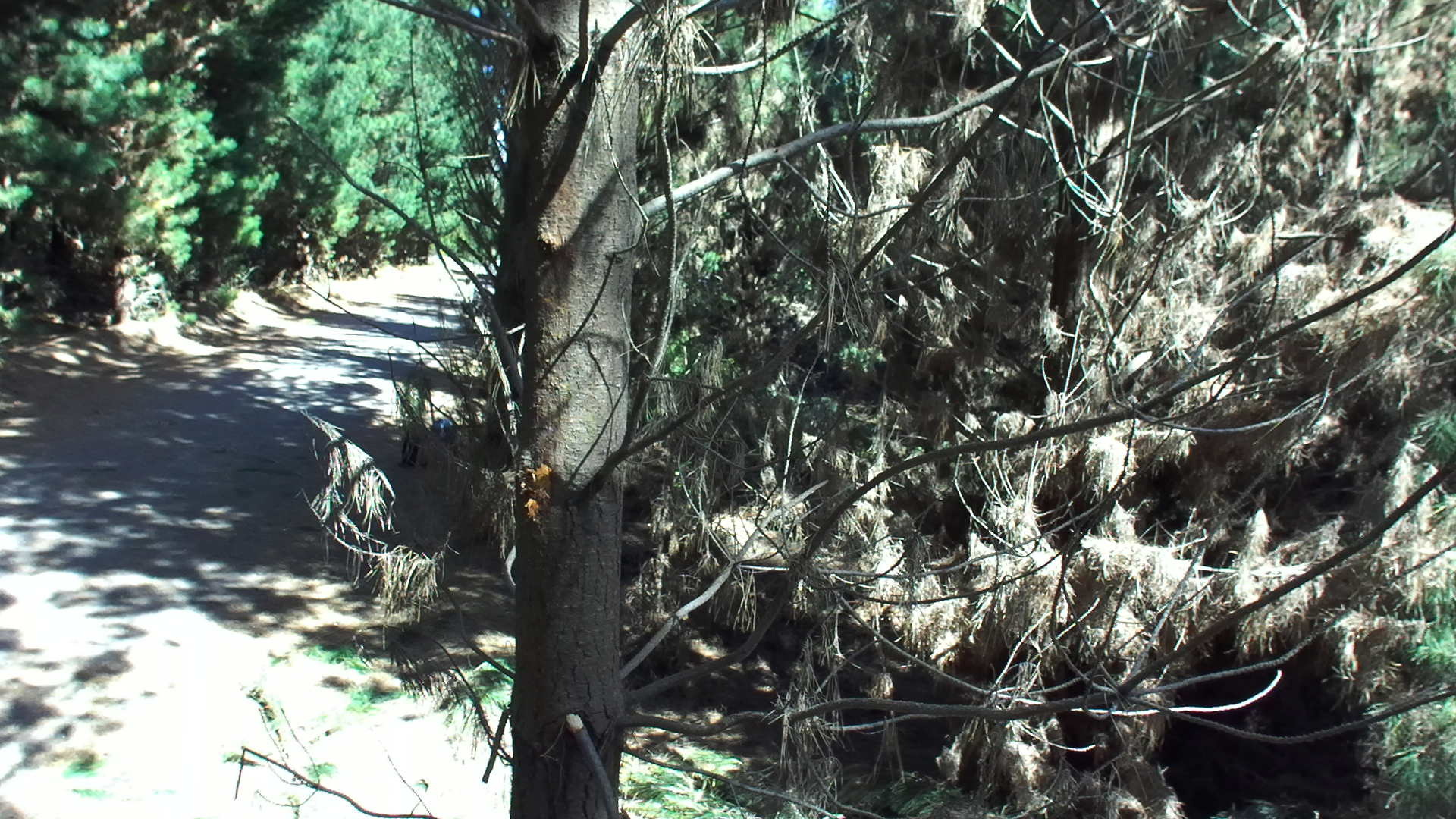}
            \caption{Left image}
        \end{subfigure}
        \hfill
        \begin{subfigure}
            [b]{0.24\textwidth}
            \includegraphics[width=\textwidth]{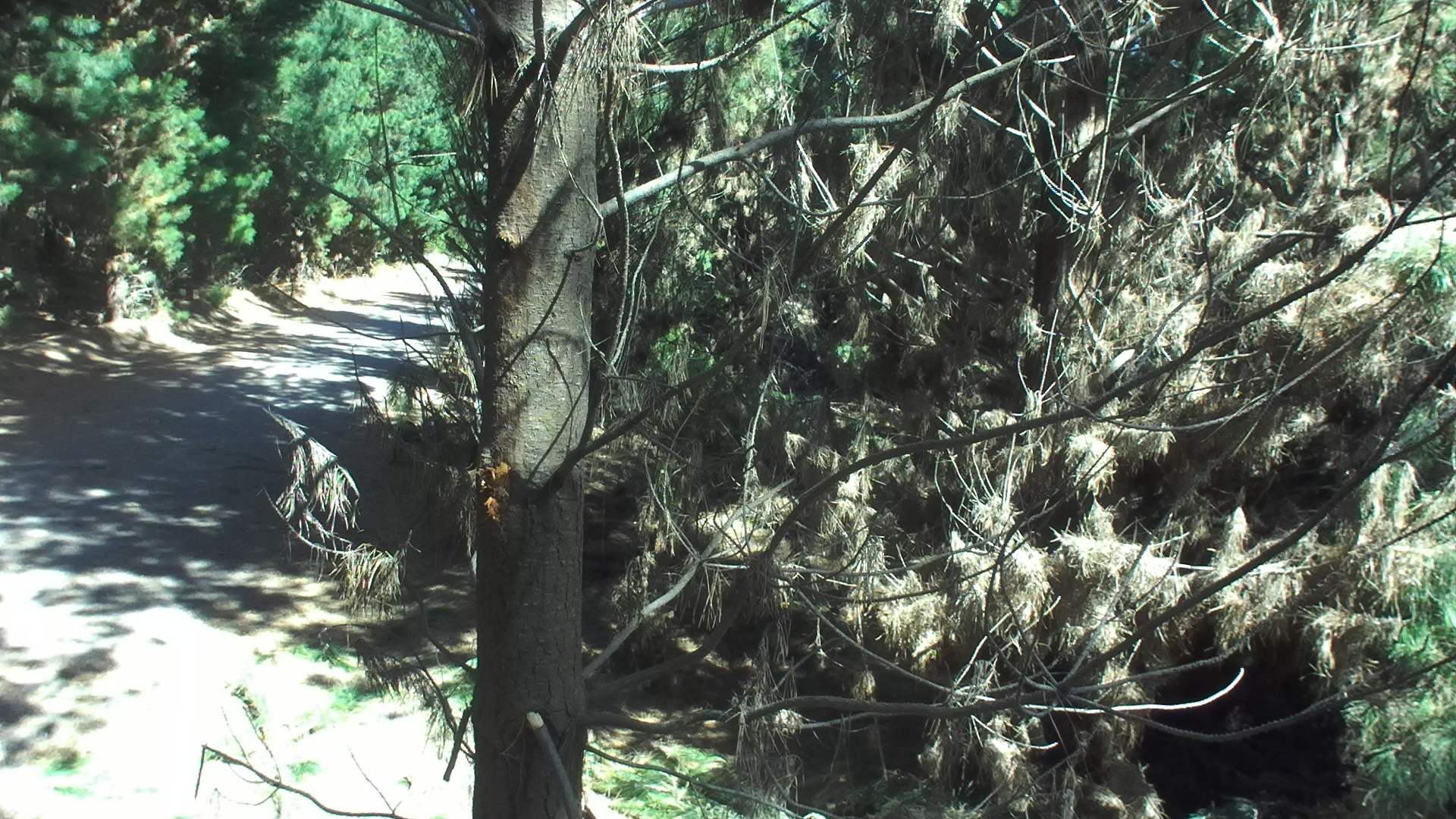}
            \caption{Right image}
        \end{subfigure}
        \hfill
        \begin{subfigure}
            [b]{0.24\textwidth}
            \includegraphics[width=\textwidth]{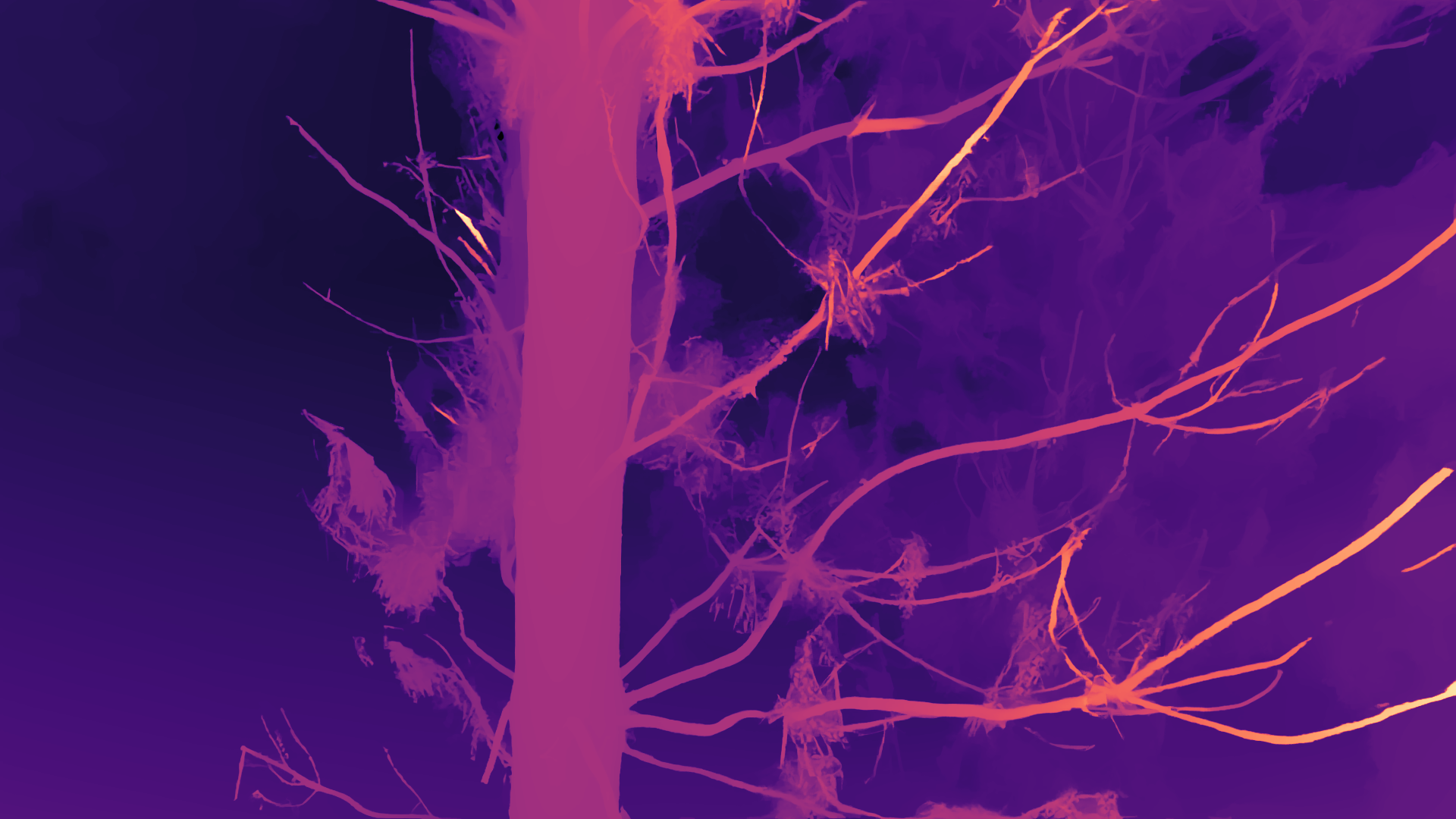}
            \caption{DEFOM prediction}
        \end{subfigure}
        \hfill
        \begin{subfigure}
            [b]{0.24\textwidth}
            \includegraphics[width=\textwidth]{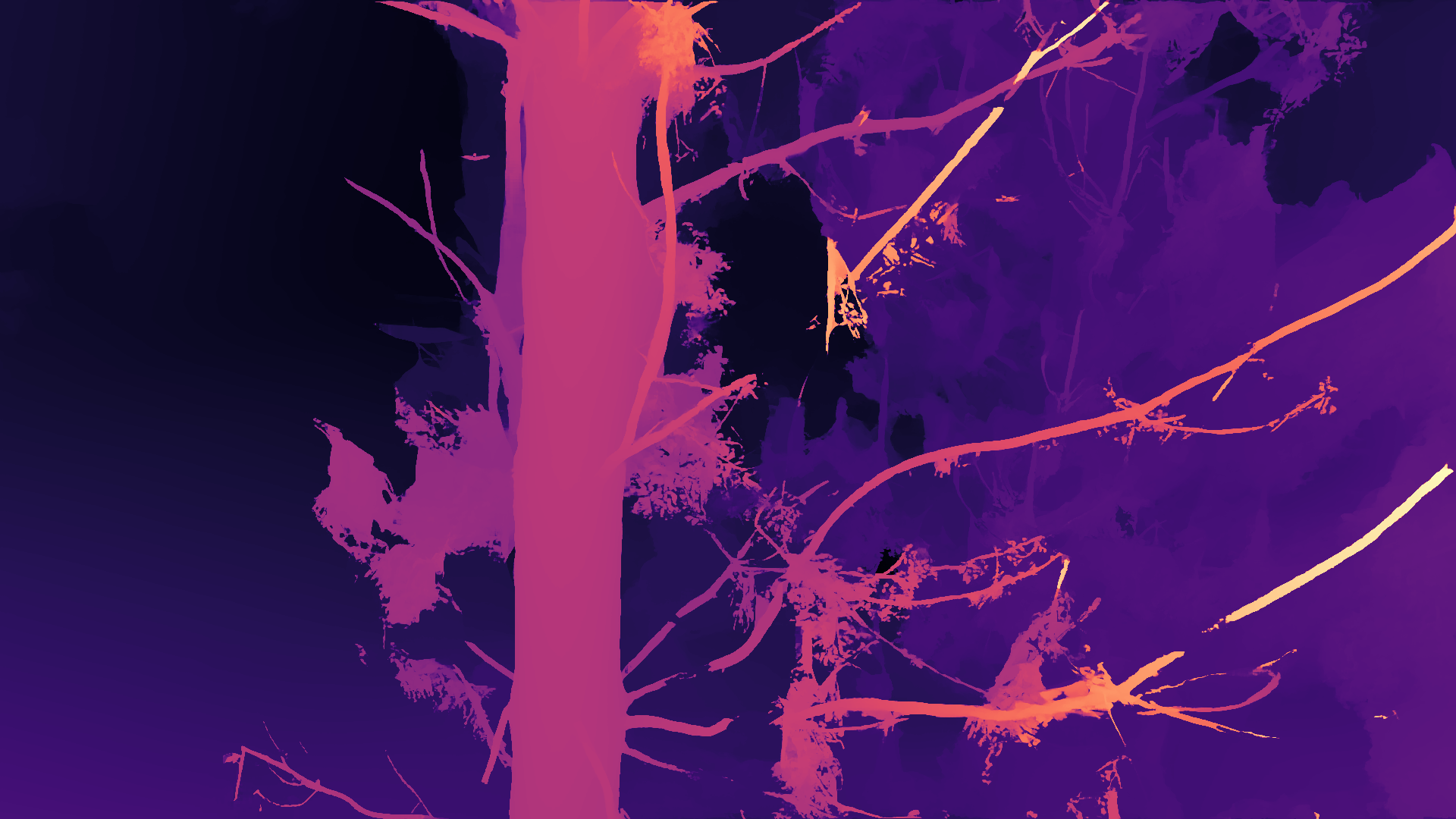}
            \caption{IGEV++ prediction}
        \end{subfigure}
        \caption{Scene 5128: Challenging dappled lighting}
        \label{fig:scene_5128}
    \end{figure*}

    \textbf{Scene 5128} (Fig.~\ref{fig:scene_5128}): Dappled illumination through
    canopy. DEFOM maintains stable, smooth predictions across shadow boundaries,
    leveraging monocular priors trained on diverse lighting. IGEV++'s iterative refinement
    produces detailed texture but less consistent depth in challenging illumination.

    \textbf{Gold-Standard Rationale}: While IGEV++ excels at detail preservation
    (beneficial for detection tasks), DEFOM's superior smoothness, consistency,
    and artifact-free predictions make it optimal for establishing reference
    depth maps. Foundation model pretraining provides robust geometric priors that
    generalize well to vegetation scenes absent from Scene Flow training. DEFOM
    predictions will serve as pseudo-ground-truth for future Canterbury dataset
    benchmarking and algorithm development.

    \section{Discussion}

    \subsection{Cross-Domain Generalization Analysis}

    Foundation models (BridgeDepth, DEFOM) demonstrate superior cross-domain
    consistency through large-scale pretraining (ETH3D: 0.23-0.35 px, KITTI:
    0.83-1.07 px, Middlebury: 4.65 px). \textbf{DEFOM achieves best ranked
    consistency} (average rank 1.75 across 4 benchmarks) with lowest coefficient
    of variation (CV: 0.58), making it ideal for establishing reference baselines
    across diverse scenes. BridgeDepth achieves lowest absolute error on ETH3D (0.23
    px) but shows higher variance across domains (CV: 0.73, rank: 2.00).

    \textbf{DEFOM vs. IGEV++ Trade-offs}: DEFOM excels in smoothness and consistency—essential
    for ground truth generation—achieving 31\% lower EPE on Middlebury (4.648 vs.
    6.775 px) and superior cross-benchmark stability. IGEV++ offers better
    outlier control (D1: 7.82\% vs. 8.28\%, 95th percentile: 18.3 vs. 22.9 px, skewness:
    1.23 vs. 1.87), valuable for safety-critical detection but less suitable as reference
    baseline due to noise in homogeneous regions.

    \textbf{Critical Validation Finding}: RAFT-Stereo's catastrophic ETH3D failure
    (98\% D1) versus normal KITTI performance (4.4-5.1\% D1) demonstrates that single-benchmark
    evaluation masks method vulnerabilities. Multi-dataset validation is
    mandatory before deployment.

    \subsection{Deployment Recommendations}

    \textbf{Canterbury Dataset Baseline}: DEFOM selected as gold-standard for pseudo-ground-truth
    generation due to optimal balance of smoothness, cross-domain consistency (rank
    1.75), and artifact-free predictions. Future algorithm development should
    benchmark against DEFOM outputs.

    \textbf{Application-Specific Selection}: Indoor/structured scenes: BridgeDepth
    (0.23 px ETH3D). Automotive: DEFOM (0.83-1.07 px KITTI, stable across conditions).
    Safety-critical detection: IGEV++ (superior outlier control: 7.82\% D1). Resource-constrained
    platforms: StereoAnywhere (zero-shot without fine-tuning).

    \subsection{Limitations and Future Work}

    \textbf{Limitations}: (1) Canterbury dataset evaluation is qualitative;
    DEFOM outputs will serve as pseudo-ground-truth for quantitative
    benchmarking. (2) Single-dataset training (Scene Flow only) limits generalization
    scope. (3) Missing methods (CREStereo, FoundationStereo unavailable at
    evaluation time).

    \textbf{Future Directions}: (1) Establish Canterbury as public benchmark with
    DEFOM pseudo-ground-truth. (2) Computational profiling for embedded UAV
    deployment. (3) Uncertainty quantification for safety-critical applications.

    \section{Conclusion}

    We present the first systematic zero-shot evaluation of eight stereo methods
    across standard benchmarks plus a novel 5,313-pair Canterbury forestry
    dataset, establishing DEFOM as the gold-standard baseline for vegetation depth
    estimation.

    \textbf{Key Contributions}: (1) \textit{Canterbury Dataset}: First UAV forestry
    stereo benchmark with vegetation-specific challenges (thin branches, occlusions,
    lighting variations). (2) \textit{Gold-Standard Baseline}: DEFOM selected
    for pseudo-ground-truth generation based on superior smoothness, cross-domain
    consistency (rank 1.75 across 4 benchmarks, CV 0.58), and artifact-free predictions.
    (3) \textit{Comparative Analysis}: IGEV++ offers better outlier control (D1 7.82\%
    vs. 8.28\%) but exhibits noise in homogeneous regions, making DEFOM more
    suitable as reference baseline despite IGEV++'s value for safety-critical
    detection.

    \textbf{Impact}: DEFOM outputs will serve as public pseudo-ground-truth for
    Canterbury dataset, enabling quantitative benchmarking of future algorithms on
    real-world forestry scenes. Multi-benchmark validation (ETH3D, KITTI, Middlebury,
    Canterbury) provides comprehensive assessment framework for UAV deployment.

    \section*{Acknowledgments}

    This research was supported by the Royal Society of New Zealand Marsden Fund
    and the Ministry of Business, Innovation and Employment. We thank the
    forestry research stations for data collection access and our annotators for
    ground truth generation.

    \bibliographystyle{IEEEtran}
    
\end{document}